\documentclass{article} % For LaTeX2e
\usepackage{iclr2026_conference,times}

\usepackage{amsmath,amsfonts,bm}

\def\eqref#1{equation~\ref{#1}}
\def\1{\bm{1}}

\DeclareMathAlphabet{\mathsfit}{\encodingdefault}{\sfdefault}{m}{sl}
\SetMathAlphabet{\mathsfit}{bold}{\encodingdefault}{\sfdefault}{bx}{n}

\usepackage[dvipsnames]{xcolor}
\usepackage[colorlinks=true,
            linkcolor=blue,
            citecolor=Rhodamine,
            urlcolor=blue]{hyperref}
\usepackage{url}
\usepackage{amsmath}
\usepackage{amssymb}
\usepackage{mathtools}
\usepackage{amsthm}
\usepackage[utf8]{inputenc} % allow utf-8 input
\usepackage[T1]{fontenc}    	% use 8-bit T1 fonts
\usepackage{url}            		% simple URL typesetting
\usepackage{booktabs}       	% professional-quality tables
\usepackage{amsfonts}       	% blackboard math symbols
\usepackage{nicefrac}       	% compact symbols for 1/2, etc.
\usepackage{microtype}      	% microtypography
\usepackage{xcolor}         		% colors
\usepackage{amsmath}
\usepackage{mathrsfs}
\usepackage{amssymb}
\usepackage{graphicx}
\usepackage{algorithm}
\usepackage{makecell}
\usepackage{lipsum}
\usepackage{enumitem}
\usepackage{graphicx}
\usepackage{listings}
\usepackage{tcolorbox}
\usepackage{multirow}
 \usepackage{algorithm}
 \usepackage{algpseudocode}
 \usepackage{booktabs}
\usepackage{multirow}
\usepackage{graphicx}
\usepackage[capitalise]{cleveref}
\crefname{equation}{Eq.}{Eqs.}
\usepackage{listings}

\title{Training Numerical Intelligence \\ via Auto-Diagnosis and Skill Discovery}

\author{Peter Chen\\
BAIR \& EECS \\
University of California, Berkeley\\
Berkeley, CA, USA \\
\texttt{pllc@eecs.berkeley.edu} \\
\And Wotao Yin \\
Decision Intelligence Lab  \\
DAMO Academy, Alibaba Group U.S. \\
Bellevue, WA, USA \\
\texttt{wotao.yin@alibaba-inc.com}
}

\iclrfinalcopy

\newcommand{\stagebox}[1]{%
  \fbox{%
    \parbox[c][2.7em][c]{7.2em}{%
      \centering #1%
    }%
  }%
}
\newcommand{\stageboxx}[1]{%
  \fbox{%
    \parbox[c][2.7em][c]{6.2em}{%
      \centering #1%
    }%
  }%
}
\newcommand{\stageboxxx}[1]{%
  \fbox{%
    \parbox[c][2.7em][c]{5.3em}{%
      \centering #1%
    }%
  }%
}
\begin{document}

\maketitle

\begin{abstract}
  AI agents are becoming increasingly capable of generating scientific code, but generating code is not the same as improving the algorithms behind it. For numerical solvers, execution feedback can expose poor performance, but rarely reveals its underlying cause and how to address it. We introduce \textbf{A}uto-\textbf{D}iagnosis and \textbf{S}kill \textbf{D}iscovery (ADSD), a framework that links numerical diagnosis to reusable solver self-improvement. ADSD follows a diagnosis-first paradigm that first explains \emph{why} a solver performs poorly, then uses this diagnosis to guide the discovery of appropriate numerical methods. The resulting knowledge is packaged into reusable solver skills, turning solver improvement from trial-and-error editing into a structured process of diagnosis, discovery, and implementation. Across four challenging numerical domains---power flow equation, AC optimal power flow control, stiff ordinary differential equations, and heterogeneous diffusion PDEs---ADSD consistently improves solver accuracy, robustness, and efficiency. On GOC-500 power flow, for example, ADSD reduces mean solver error by nearly $71\times$, with improvements further transferring to unseen grid topologies and operating regimes.
\end{abstract}

%see intro-long.tex for long version

\section{Introduction}\label{sec:intro}

Large language models (LLMs) are becoming increasingly capable of writing scientific code and discovering algorithms~\citep{romera2024mathematical, novikov2025alphaevolve}. However, generating a numerical solver is fundamentally different from building standard software. A solver may run without errors and produce seemingly reasonable outputs, yet remain unnecessarily slow or fail on challenging instances due to numerical issues such as ill-conditioning, stiffness, or repeated matrix factorizations. Addressing these failures requires more than conventional debugging. An AI agent must understand \emph{why} the solver behaves poorly, diagnose the underlying numerical condition, and select a numerical method or modification tailored to that condition.

This need for numerical diagnosis exposes a major blind spot in today's scientific agents. Existing systems can generate or revise numerical code using execution traces and error feedback \citep{gaonkar2025sciml, du2026autonumerics}, search the literature for relevant techniques \citep{garikaparthi2025mir, asai2024openscholar}, and convert research papers into reusable tools \citep{gandhi2025researchcodeagent, miao2025paper2agent, shen2026skillfoundry}. However, these capabilities are rarely coordinated. Execution feedback often leads to trial-and-error edits, while literature search relies on broad keywords. As a result, an agent may detect poor solver performance without identifying its cause or the methods best suited to address it. This challenge is particularly notorious in numerical computing, where the same symptom can arise from different underlying causes. Since numerical methods also rely on specific assumptions and trade-offs, effective improvement requires first diagnosing the underlying numerical pathology and then identifying methods whose mechanisms and assumptions match it.

Motivated by this gap, we argue that an effective numerical intelligence should explicitly separate three stages that current workflows often blur together: (i) \textbf{observable behavior}, which captures raw runtime evidence, such as residual histories and convergence logs; (ii) \textbf{diagnostic knowledge}, which interprets these observations by forming testable links between solver symptoms and their underlying numerical conditions; and (iii) \textbf{executable solver knowledge}, which finally translates the diagnosis into concrete numerical algorithms that are correctly implemented and packaged for direct integration into the solver. Building on this separation, we introduce \textbf{A}uto-\textbf{D}iagnosis and \textbf{S}kill \textbf{D}iscovery (ADSD), a automatic framework without updating model weights for numerical intelligence that operationalizes these stages through a structured, diagnosis-first loop:
\[
\stageboxx{solver behavior}
\xrightarrow{\text{diagnose}}
\stagebox{verified numerical\\condition}
\xrightarrow{\text{retrieve}}
\stagebox{targeted scientific\\knowledge}
\xrightarrow{\text{implement}}
\stageboxxx{executable\\solver skill}
\]

We construct ADSD across four challenging classes of numerical problems: power flow equation solving, AC optimal power flow control, stiff ODE integration, and heterogeneous diffusion PDEs, via three leading agent baselines, including GPT-5.5, Claude-sonnet-5, and Qwen3.8-2.4T. Moreover, our evaluations stress the numerical regimes with real-world engineering pressure, such as topology and feasibility shifts in power flow, near-loadability AC-OPF edge cases, stiff dynamical systems, and ill-conditioned high-contrast diffusion problems. Across these settings, ADSD consistently improves the solvers generated by the base models while preserving or improving computational efficiency. On power flow, for example, ADSD reduces Qwen3.8-2.4T's mean error by a $71\times$ error reduction on GOC-500 power flow solver benchmark. Together, these results show that ADSD generates numerical knowledge that improves solver accuracy and efficiency across diverse scientific-computing tasks.

\vspace{-0.5em}
\paragraph{Contributions.} By introducing ADSD, we (i) provide a practical and novel diagnosis-first paradigm for numerical intelligence, enabling agents to move systematically from solver behavior to verified numerical diagnoses and targeted numerical skills, and (ii) make ADSD as a general framework for automated agent \emph{self-improvement}, demonstrating consistent improvements across diverse numerical domains, language model families, and unseen problem regimes.

\vspace{-0.5em}
\paragraph{Related works.} Recent work advances numerical intelligence from algorithm discovery and simulation agents \citep{romera2024mathematical, novikov2025alphaevolve}. In power engineering, the scale and complexity of real-world systems have led prior work to rely on multi-agent pipelines for solver construction and orchestration \citep{jia2025enhancing, jin2025gridmind, shamseldein2026gridmind}. Similar challenges arise in differential-equation solving for large dynamical systems. Prior work also studied whether LLMs can effectively use numerical solvers \citep{gaonkar2025sciml}, refine generated PDE solvers through execution feedback and debugging \citep{li2025codepde, du2026autonumerics}, and improve numerical reasoning through reinforcement learning \citep{cai2026reinforcement}. However, across both domains, existing methods provide limited mechanisms for transferring observed solver behavior to numerical deficiencies and acquiring relevant methods beyond its inherent knowledge, so a \emph{general} and \emph{efficient} pipeline for training \emph{self-evolving} numerical intelligence remains underexplored.

Beyond numerical intelligence, Voyager \citep{wang2025voyager} first demonstrated that embodied agents can acquire and reuse executable skills with the gradient-free paradigm. More recent work extends this idea to scientific settings by converting papers and other scientific resources into executable skills and workflows \citep{shen2026skillfoundry,gandhi2025researchcodeagent,miao2025paper2agent}, while using execution traces and validation feedback to guide skill discovery, refinement, and selection \citep{liu2026skillforge,moll2026grasp,alzubi2026evoskill,liu2026skillrevise,chen2026skillcat}. Complementary studies further investigate trajectory-based failure diagnosis and automated hypothesis falsification \citep{barke2026agentrx,huang2025automated}. Building on these directions, our work focuses specifically on diagnosis-guided acquisition of skills for numerical methods, and we defer discussions towards further related works to Appendix~\ref{app:rw}.
\section{Preliminaries}\label{sec:prelim}

A numerical solver computes an approximate solution to a mathematical problem given instance-specific parameters and conditions. In this paper, we study four settings for numerical solver:
\vspace{-0.5em}
\paragraph{Power flow equations.}
Power flow (PF) determines a steady-state voltage configuration that satisfies a grid's specified power injections, voltage controls, and network physics. After accounting for prescribed quantities and the reference voltage angle, the problem can be written as
\(
F_{\text{PF}}(x;\xi)=0
\), where $x$ contains the unknown voltage angles and magnitudes, $\xi$ specifies the network and operating conditions, and $F_{\text{PF}}$ collects the active- and reactive-power mismatches. A solver seeks a voltage state with sufficiently small residual $\|F_{\text{PF}}(x;\xi)\|_{\infty}$. This residual measures satisfaction of the power-balance equations, rather than voltage error relative to a reference solution.

\vspace{-0.5em}
\paragraph{AC optimal power flow.}
AC optimal power flow (AC-OPF) optimizes an operating objective subject to power balance and operational limits. Let $z=(v,\theta,p^g,q^g)$ collect voltage magnitudes, voltage angles, and generator outputs. The problem takes the form
\[
\min_z C(p^g)
\quad \text{s.t.} \quad
F_{\text{AC}}(z;\xi)=0, \quad g(z;\xi)\leq 0,
\]
where $C$ is the generation cost, $F_{\text{AC}}$ enforces power balance, and $g$ represents generator, voltage, and transmission limits, so the performance depends on both constraint satisfaction and objective quality.

\vspace{-0.5em}
\paragraph{ODEs.}
An ordinary differential equation (ODE) integrator approximates the trajectory of an initial-value problem (IVP),
\[
\dot{y}(t)=f(t,y(t);\eta), \qquad y(t_0)=y_0,
\]
where $\eta$ denotes system parameters. Our experiments consider systems-biology models that often exhibit stiffness, for which stability can impose restrictive step sizes on explicit methods \citep{stadter2021benchmarking}. Note that the final integration accuracy and cost depend on the numerical scheme, step-size selection, and error control, and we evaluate accuracy against reference trajectories and measure computational work through runtime and right-hand-side evaluations.

\vspace{-0.5em}
\paragraph{PDEs.}
An elliptic partial differential equation (PDE) solver approximates the spatial solution of a boundary-value problem,
\begin{equation}\label{eq:pde}
  -\nabla\cdot\bigl(k(x)\nabla u(x)\bigr)=f(x), \qquad x\in\Omega, \qquad u|_{\partial\Omega}=0,
\end{equation}
where $k(x)>0$ denotes the diffusion coefficient and $f(x)$ the forcing. We apply two-dimensional steady-state diffusion problems with heterogeneous, high-contrast coefficients from  \citet{trifonov2024condiff}. Specifically, the large coefficient contrasts can produce ill-conditioned discrete systems, slowing iterative convergence, making it challenging for diagnosis and solver improvement. 

\section{Main Results}\label{sec:mr}
\begin{figure*}[t]
    \vspace{-0.5em}
    \centering
\includegraphics[width=0.9\linewidth]{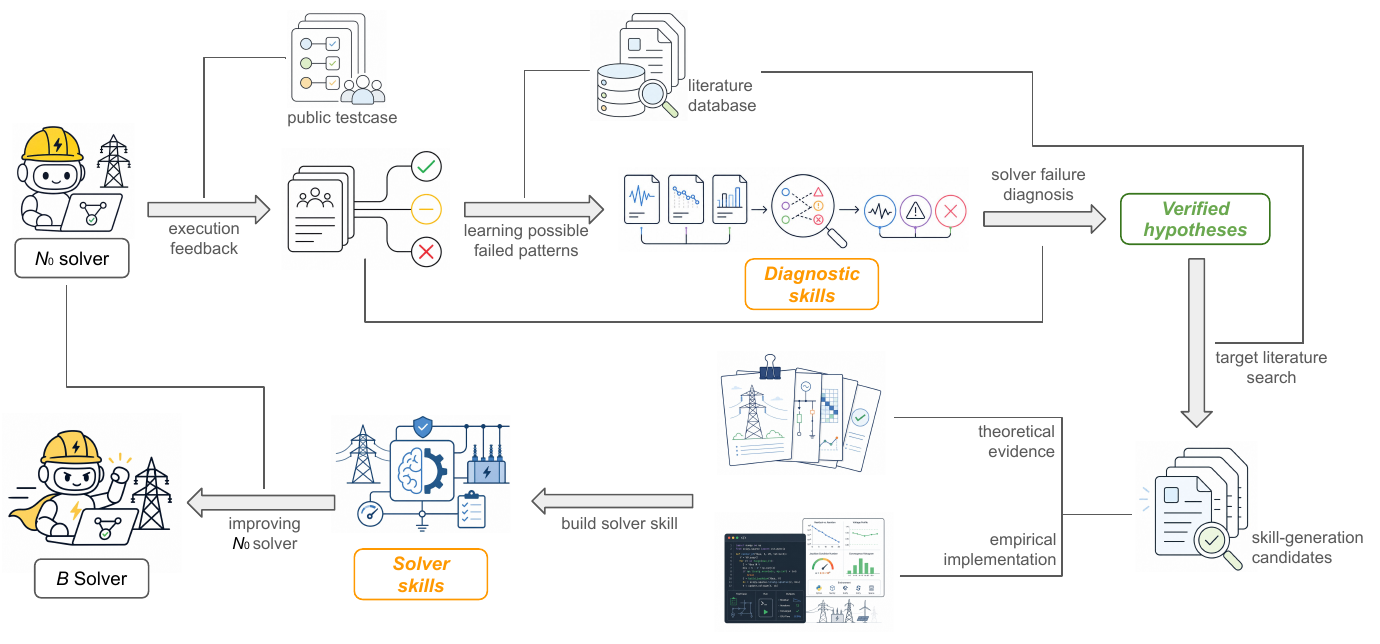}
    \vspace{-0.5em}
    \caption{Overview of ADSD. \textbf{Auto-diagnosis} combines execution feedback from the frozen baseline solver $N_0$ on public test cases with literature-derived diagnostic skills to identify numerical failure patterns. \textbf{Skill discovery} uses verified hypotheses to retrieve candidate methods and complete their implementations. The resulting reusable solver skills are supplied, together with $N_0$, to a fresh agent to generate the improved solver $B$.}
    \label{f1}
    \vspace{-1.5em}
\end{figure*}

We introduce the two core components of ADSD: auto-diagnosis and skill discovery. In practice, a numerical agent must first understand why its solver struggles (i.e., diagnosis) before it can identify the right knowledge and tools to improve it. In our pipeline, auto-diagnosis identifies numerical conditions that explain a solver's behavior, while skill discovery finds and builds re-usable empirical and theoretical methods from the literature. Together, they form a diagnosis-driven pipeline that turns observed solver behavior into evidence-grounded solver improvements. Before running the pipeline, we partition the test cases into \emph{public} and \emph{private} sets. Feedback from the public cases is available for training and development, whereas the private cases remain hidden throughout the process and are reserved for final evaluation. A literature database across power-grid and dynamic system solvers is also pre-constructed, serving as the specialized knowledge source for diagnosis and skill discovery.

\subsection{Auto-diagnosis}\label{sec:AD}

\begin{algorithm}[!h]
\caption{Auto-diagnosis pipeline}
\label{alg:auto-diagnosis}
\begin{algorithmic}[1]
\Require Public testcases $\mathcal{P}$, literature database $\mathcal{L}$, solver contract
\Ensure Diagnostic skills $\mathcal{S}$ and verified hypotheses $\mathcal{H}$
\State Generate and freeze the baseline solver $N_0$
\State $\mathcal{O} \gets$ run $N_0$ once on $\mathcal{P}$ and record observable behavior

\State $\mathcal{L}' \gets$ filter relevant papers from $\mathcal{L}$
\State $\mathcal{C} \gets$ read shortlisted papers and extract page-supported claims
\State $\mathcal{S} \gets$ cluster claims $\mathcal{C}$ by shared computation and build diagnostic probes

\State Split $\mathcal{O}$ into diagnosis rounds \(\{\mathcal{D}_1,\mathcal{D}_2,\cdots,\mathcal{D}_n\}\) and a verification pool $\mathcal{V}$
\State $\mathcal{H} \gets \emptyset$
\ForAll{diagnosis rounds $\mathcal{D}_i$}
    \State Execute $\mathcal{S}$ on $\mathcal{D}_i$ and propose diagnostic hypotheses 
    \State Verify the proposed hypotheses on $\mathcal{V}$
    \State Add successfully verified hypotheses to $\mathcal{H}$
\EndFor

\State \Return $\mathcal{S}, \mathcal{H}$
\end{algorithmic}
\end{algorithm}

Our auto-diagnosis framework can be divided into the three key steps: (i) \textbf{baseline solver generation}, where we freeze a baseline solver and collect its observable solving behavior on public testcases; (ii) \textbf{diagnostic skill construction}, where literature-grounded numerical knowledge is converted into executable diagnostic tests; and (iii) \textbf{hypothesis verification}, where the probes are used to identify and validate relationships between solver behavior and underlying numerical conditions.

As shown in Algorithm~\ref{alg:auto-diagnosis}, after data processing, the agent generates a baseline solver $N_0$ using its inherent knowledge and capabilities, with access to the public testcases for execution and debugging, as in standard coding-agent settings. Once finalized, $N_0$ is frozen and rerun on all public testcases to record its observable behavior, including quantitative trajectories and summary statistics, which form the empirical basis for subsequent diagnosis. Throughout the full ADSD pipeline, the agent is prohibited from accessing off-the-shelf solvers, since our goal is to diagnose and improve the numerical capabilities already possessed by the base agent; off-the-shelf solver implementations would confound this comparison.

\vspace{-0.5em}
\paragraph{Symptoms-condition diagnosis.} The key here is to connect observable symptoms to the underlying numerical conditions that may cause them. Corresponding to lines 3-5 of Algorithm~\ref{alg:auto-diagnosis}, we construct diagnostic skills via three steps. The agent first screens each paper using a pre-processed preview containing its metadata, abstract, introduction, conclusion from the main text, retaining papers likely to contain concrete symptom-condition relationships. Shortlisted papers are then read in full, and each relationship is recorded as a diagnostic claim with exact evidence, together with the observable quantities and running statistics from $N_0$. Finally, claims sharing a common diagnostic computation are grouped into diagnostic skills for next stage to generate and verify possible hypotheses.

\vspace{-0.5em}
\paragraph{Hypothesis discovery.} We use the resulting skill set $\mathcal{S}$ to explain recurring solver behavior. First, we randomly seperate the public observations $\mathcal{O}$ into non-overlapping diagnosis rounds $\{\mathcal{D}_1,\ldots,\mathcal{D}_n\}$ and a verification pool $\mathcal{V}$, which checks the effectiveness of the hypotheses produced from each diagnosis round $\mathcal{D}_{i}$. Specifically, we perform diagnosis over multiple rounds to keep each round small enough for the agent to inspect individual traces and less frequent behaviors. For example, on IEEE-118, we split 5,000 public observations into 1,000 diagnosis cases across five rounds of 200 cases each, together with 4,000 verification cases. For each round $i$, the agent apply every skill in $\mathcal{S}$ to the cases in $\mathcal{D}_i$. Using the diagnostic skills to examine each testcase's input, solver trace, and final output, the agent then proposes hypotheses that connect observed solver symptoms to suspected numerical conditions. Each hypothesis is then converted into a deterministic \emph{detector} with explicit criteria for when the hypothesized relationship is considered supported or falsified. Each detector is then applied to $\mathcal{V}$, and hypotheses that passed over $90\%$ of $\mathcal{V}$ are added to the verified set $\mathcal{H}$ and marked as a common ``illness'', while rejected hypotheses are discarded. Finally, each verified hypothesis becomes a targeted search guidance for skill discovery, and ADSD then finds the literature for methods that explain or address the common illness and converts sufficiently supported methods into executable solver skills to tackle such illness.

\subsection{Skill discovery}\label{sec:SD}

Our skill-discovery framework converts verified diagnostic hypotheses into executable solver-improvement skills through four stages: (i) \textbf{hypothesis-guided paper screening}, which identifies methods relevant to the diagnosed numerical condition; (ii) \textbf{theoretical claim verification}, which checks the paper's claims, assumptions, and compatibility with $N_0$; (iii) \textbf{candidate construction}, which empirically implements the methods; and (iv) \textbf{skill construction}, which packages verified methods into skills for generating an improved solver.

Unlike diagnostic screening, which searches broadly for measurable symptom-condition relationships, we design this stage to specifically target reproducible numerical methods that may address any verified hypothesis. For each paper, the agent first receives a similar bounded preview to evaluate its mechanism relevance and compatibility with $N_0$, yielding a filtered candidate set for each hypothesis.

Candidate papers are then read in full to determine whether their methods are scientifically supported and transferable to $N_0$. For each relevant claim, the agent records exact page-level evidence, including equations, algorithms, and assumptions (we present these details for later case studies in Appendix~\ref{app:case}). A separate theory audit rechecks the derivation and evaluates whether its assumptions are compatible with the $N_0$ solver. Each supported method is then summarized into a \emph{theoretical claim dossier}, which serves as the basis for its empirical implementation (e.g., actual codes). The resulting implementation is empirically evaluated for reproduction of the claimed results, and verified candidates are packaged as frozen solver skills containing the executable implementation, usage instructions, and claim dossier. To generate the improved solver, we provide the original $N_0$ solver and resulting skill packets to the agent using the same model, reasoning budget, runtime, and original solver-generation prompt as for $N_0$. The agent integrates the supplied skills to produce a new solver $B$.

\section{Experiments}\label{sec:exp}

Our experiments consist of four main scanerios, together with additional results over further ablation analyses in \S\ref{sec:aa}. The four main experiments study power-flow equation solving (\S\ref{sec:exp-pf}), AC optimal power-flow control and optimization (\S\ref{sec:acopf}), ODE integration (\S\ref{sec:ode-int}), and 2D diffusion PDE solving (\S\ref{sec:pde}). Across all four benchmarks, we select recent datasets that represent challenging numerical tasks and closely reflect the practical difficulties faced by human engineers.

\begin{table*}[t]
\centering
\caption{Performance on the IEEE-118 and GOC-500 validation sets. Values are mean $\pm$ standard deviation. ``$N_0$ solver'' stands for the baseline performance using agent's inherent knowledge, and ``+ ADSD'' denotes the agent's performance after training with ADSD pipeline.}
\label{tab:main-results}
\small
\setlength{\tabcolsep}{8pt}
\resizebox{\linewidth}{!}{
\begin{tabular}{llcccc}
\toprule
\multirow{2}{*}{\textbf{Base Model}}
& \multirow{2}{*}{\textbf{Configuration}}
& \multicolumn{2}{c}{\textbf{IEEE-118}}
& \multicolumn{2}{c}{\textbf{GOC-500}} \\
\cmidrule(lr){3-4}\cmidrule(lr){5-6}
& & \textbf{Run time} $(\downarrow)$
& \textbf{Error} $(\downarrow)$
& \textbf{Run time} $(\downarrow)$
& \textbf{Error} $(\downarrow)$ \\
\midrule

\multirow{2}{*}{\texttt{GPT-5.5}}
& $N_0$ solver
& $0.73 \pm 0.21$
& $9.1{\times}10^{-10} \pm 2.0{\times}10^{-9}$
& $0.51 \pm 0.14$
& $9.3{\times}10^{-10} \pm 2.2{\times}10^{-9}$ \\

& \quad + ADSD
& $0.55 \pm 0.15$
& $2.7{\times}10^{-10} \pm 9.9{\times}10^{-10}$
& $0.50 \pm 0.14$
& $1.6{\times}10^{-10} \pm 7.4{\times}10^{-10}$ \\

\midrule

\multirow{2}{*}{\texttt{Claude-Sonnet-5}}
& $N_0$ solver
& $0.16 \pm 0.04$
& $1.1{\times}10^{-9} \pm 2.1{\times}10^{-9}$
& $0.52\pm 0.26$ & 
$7.9{\times}10^{-10} \pm 1.9{\times}10^{-9}$ \\

& \quad + ADSD
& $0.15 \pm 0.04$
& $7.4{\times}10^{-10} \pm 1.8{\times}10^{-9}$
& $0.52\pm 0.25$ & 
$1.4{\times}10^{-10} \pm 8.4{\times}10^{-10}$\\
\midrule

\multirow{2}{*}{\texttt{Qwen3.8-2.4T}}
& $N_0$ solver
& $0.57 \pm 0.02$
& $9.9{\times}10^{-10} \pm 2.1{\times}10^{-9}$
& $0.18\pm0.09$  & $1.0{\times}10^{-9}\pm2.1{\times}10^{-9}$ \\

& \quad + ADSD
& $0.54\pm 0.01$
& $6.5{\times}10^{-10} \pm 1.6{\times}10^{-9}$ 
& $0.18\pm0.10$  &  $1.4{\times}10^{-11}\pm2.4{\times}10^{-10}$
\\
\bottomrule
\vspace{-2em}
\end{tabular}}
\end{table*}

\subsection{Power flow  (PF) equation solver}\label{sec:exp-pf}

\paragraph{Benchmark data.} We apply training and evaluation cases from the IEEE and GOC grid families in PF$\Delta$ \citep{rivera2025pf}, which provides generalization settings along multiple dimensions. First, it includes multiple grid topologies: \texttt{N} retains the full network, whereas \texttt{N1} and \texttt{N2} contain one and two fewer grid components, respectively. Moreover, it covers different operating regimes, including feasible (\texttt{F}) and close-to-infeasible (\texttt{IF}) cases. These regimes move progressively closer to the steady-state voltage-stability boundary and therefore become increasingly numerically challenging.

\vspace{-0.5em}
\paragraph{Training \& evaluation setup.} We conduct training and evaluation on both the IEEE and GOC benchmark families. Because these families use different problem formulations and input-output schemas, a solver developed for the IEEE setting cannot be directly applied to GOC. We consider three evaluation settings: (i) \textbf{In-domain evaluation}, which uses the same system configuration for training and evaluation but evaluates on a disjoint validation set; for example, the solver is trained on the IEEE-118 training set and evaluated on the same type of IEEE-118 validation set. (ii) \textbf{Topology generalization}, which evaluates the solver under a different topology distribution; for example, the solver is trained on IEEE-118-\texttt{N2} and evaluated on IEEE-118-\texttt{N1}. (iii) \textbf{Feasibility generalization}, which evaluates the solver under more harsh feasibility, e.g., the solver is trained on IEEE-118-\texttt{F} and evaluated on IEEE-118-\texttt{IF}. For the agent baselines, we select recent high-performing models that offer a practical balance between agentic capability and inference cost, including closed-source model, GPT-5.5 and Claude-Sonnet-5, and one of strongest open-source models, Qwen3.8-2.4T.

\vspace{-0.5em}
\paragraph{In-domain evaluation.} We train the ADSD pipeline on the public test cases following the procedure in \S\ref{sec:mr}, and then evaluate the resulting solvers on a hidden validation set drawn from the same grid topology and operating-regime difficulty. For example, the IEEE-118 benchmark contains 5,000 public cases and 10,000 hidden validation cases, with both sets using the \texttt{N2} topology configuration and the feasible operating regime \texttt{F}. Table~\ref{tab:main-results} reports solver wall-clock time and error on the IEEE-118 and GOC-500 validation sets. Across the evaluated base models, ADSD consistently improves solver performance, with the most significant gain observed for the Qwen3.8 baseline on GOC-500, where the agent reduces solver error by approximately $71\times$ while maintaining comparable solver speed. These results indicate that ADSD can transform numerical patterns discovered from public cases into reusable solver skills that improve solution accuracy while preserving computational efficiency.
\begin{figure*}[t]
    \vspace{-1em}
    \centering
\includegraphics[width=0.98\linewidth]{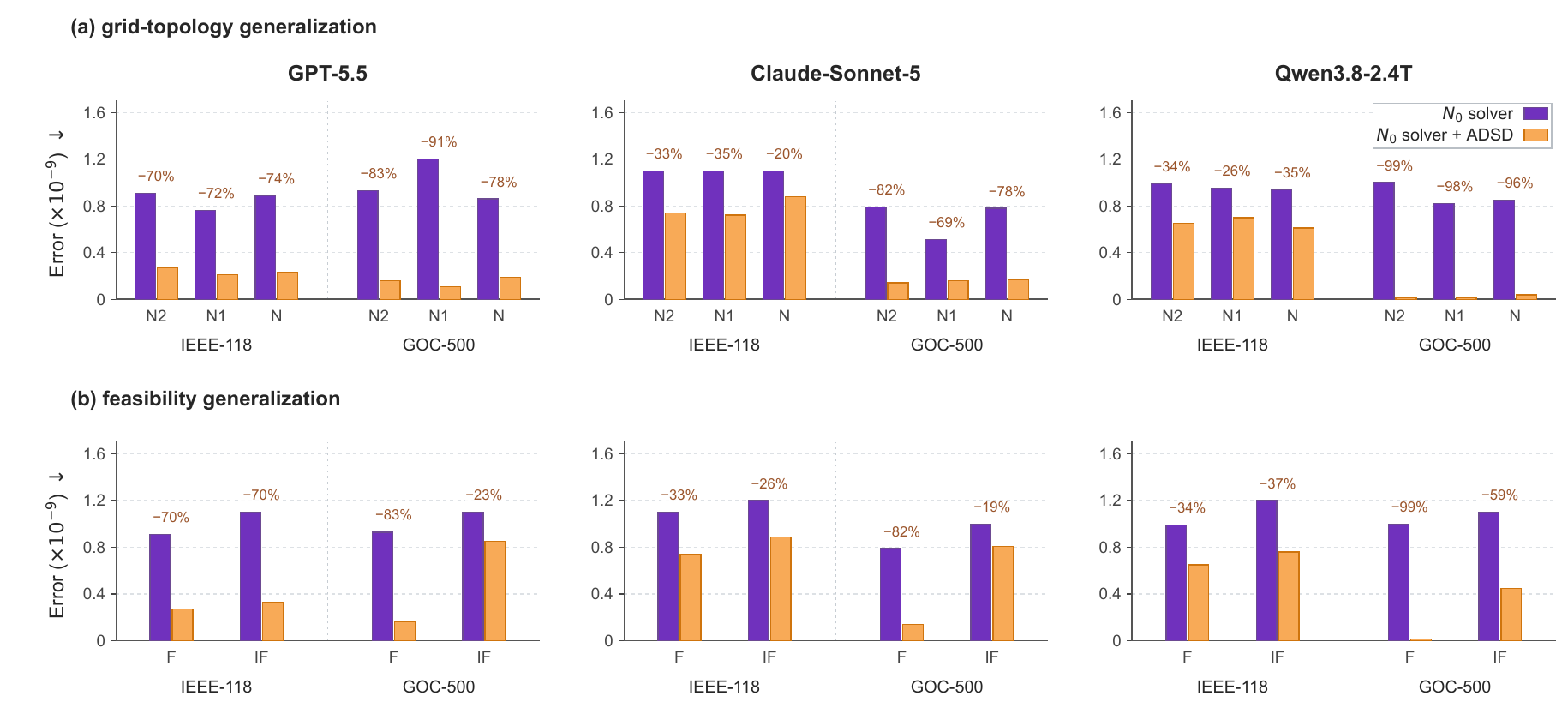}
    \vspace{-1em}
    \caption{ADSD performance on IEEE-118 and GOC-500 under (a) topology generalization from \texttt{N2} to unseen \texttt{N1}/\texttt{N} grids and (b) feasibility generalization from feasible (\texttt{F}) to close-to-infeasible (\texttt{IF}) cases. Over all cases, learned skills can be well generalized to different grid setups and difficulties. Detailed results for figures (a) and (b) are presented in Tables~\ref{tab:topology-generalization} and \ref{tab:feasibility-generalization}, respectively.}
    \label{f2}
    \vspace{-1.5em}
\end{figure*}

\vspace{-0.5em}
\paragraph{Grid-topology generalization.}
We next examine whether solver skills learned exclusively from \texttt{N2} topology cases can generalize without further adaptation to structurally more complete networks: \texttt{N1} cases with one component removed and \texttt{N} cases with the full grid intact. As shown in Figure~\ref{f2}(a), the skills learned from \texttt{N2} generalize well across these unseen topologies, with significant improvement in solver performance across all base models and grid benchmarks. We note that this capability is particularly important in real power-grid applications, where engineers must routinely handle multiple network topologies, making it desirable for skills learned from one topology to remain reusable across different grid configurations.

\vspace{-0.5em}
\paragraph{Grid-feasibility generalization.} We further evaluate feasibility generalization by transferring skills learned exclusively from feasible cases (\texttt{F}) to substantially more challenging close-to-infeasible cases (\texttt{IF}). While feasible cases admit valid operating solutions, close-to-infeasible cases lie near at grid constraints, where ill-conditioning can make power-flow equations difficult to solve. This creates a substantially more demanding setting that better reflects difficult scenarios encountered by power-system engineers. The corresponding results are shown in Figure~\ref{f2}(b). Unlike topology generalization, the room for improvement on close-to-infeasible cases is naturally more limited because the underlying grid constraints must be satisfied near the voltage-stability boundary. Nevertheless, even under these limitations, the learned skills still improve the infeasible-case solvers, demonstrating that the acquired numerical knowledge remains useful beyond the feasible training regime.

\paragraph{Case study.} To help readers better understand ADSD, we take GPT baseline over PF$\Delta$ setup and Qwen baseline with $84\%$ pass rate improvement from AC-OPF setup, presenting a detailed technical case study in Appendix~\ref{app:case}, examining how each skill assists the agent to build the better solver.

\vspace{-0.5em}
\subsection{AC optimal power flow (AC-OPF) solver} \label{sec:acopf}
Different from the more recent PF$\Delta$ benchmark, which focuses on solving power-system equations and evaluating generalization ability, OPFData \citep{lovett2024opfdata} targets classical power-system control and optimization. This extends our evaluation beyond solving AC power-flow equations to optimizing generation cost subject to the same physical equations and operational constraints. Such optimization becomes especially challenging as the system approaches its operating limits, motivating our evaluation on edge cases that resembles the realistic extreme real-world engineering difficulty.

\vspace{-0.5em}
\paragraph{Edge-case design.} Power-system operation becomes particularly challenging under high electricity demand, when operators must continue serving load at reasonable cost without violating grid operating limits. In these stressed regimes, generation and voltages are tightly coupled: adjusting generation at one location can alter network conditions elsewhere, so resolving one constraint may introduce another. Reliable optimization is therefore most critical near the system's operating boundary, where feasible margins are small and the solver must make effective decisions under severe network stress.

\vspace{-0.5em}
\paragraph{Training \& evaluation setup.} To capture this realistic difficulty, among 300,000 cases from the IEEE-57 grid environment, we specifically filter 7,000 edge cases (2,000 for training and 5,000 for hidden evaluation) that operate near the maximum loadability of the system, with simultaneously high demand and transmission congestion. We evaluate solver quality using mean absolute relative objective error, $|f-f_{\text{ref}}|/\max(|f_{\text{ref}}|,1)$, where $f_{\text{ref}}$ denotes the reference result from the benchmark, together with runtime and pass rate. As shown in Table~\ref{tab:acopf-main-results}, ADSD improves solver robustness and efficiency under these highly stressed operating conditions while maintaining low objective error.

\begin{table*}[t]
  \centering
  \caption{Performance on 5,000 AC-OPF hidden evaluation cases. $N_0$ denotes the original solver and ADSD denotes its skill-augmented solver. Mean error is calculated as $|f-f_{\mathrm{ref}}|/\max(|f_{\mathrm{ref}}|,1)$, across all cases that $N_0$ and ADSD solver passed in common.}
  \label{tab:acopf-main-results}
  \small
  \setlength{\tabcolsep}{8pt}
  \resizebox{\linewidth}{!}{%
  \begin{tabular}{llccc}
  \toprule
  \textbf{Base Model}
  & \textbf{Configuration}
  & \textbf{Mean error} $(\downarrow)$
  & \textbf{Runtime} $(\downarrow)$
  & \textbf{Pass rate} $(\uparrow)$ \\
  \midrule

  \multirow{2}{*}{\texttt{GPT-5.5}}
  & $N_0$ solver
  & $4.8\times10^{-7}\pm 6.5\times10^{-7}$
  & $0.80 \pm 0.22$
  & $97.82\%$ \\
  & \quad + ADSD
  & $4.0\times10^{-7}\pm 6.4\times10^{-7}$
  & $0.74 \pm 0.24$
  & $99.26\%$ \\

  \midrule
  \multirow{2}{*}{\texttt{Claude-Sonnet-5}}
  & $N_0$ solver
  & $2.5\times10^{-6}\pm 5.5\times10^{-6}$
  & $0.62 \pm 0.11$
  & $78.78\%$ \\
  & \quad + ADSD
  & $1.9\times10^{-6}\pm 5.4\times10^{-6}$
  & $0.42 \pm 0.09$
  & $81.42\%$ \\

  \midrule
  \multirow{2}{*}{\texttt{Qwen3.8-2.4T}}
  & $N_0$ solver
  & $2.7\times10^{-6}\pm8.6\times10^{-7}$ & $0.26\pm0.04$ & $14.42\%$ \\
  & \quad + ADSD
  & $1.2\times10^{-6}\pm8.2\times10^{-7}$ & $0.18\pm0.05$& $98.36\%$ \\
  \bottomrule
  \vspace{-2.5em}
  \end{tabular}%
  }
\end{table*}

\subsection{ODE integrator}\label{sec:ode-int}

We use ODE benchmark data from the systems-biology model collection of \citet{stadter2021benchmarking}. After processing, the benchmark contains 111 distinct system families. We partition the testcases from these models into 39 for training and 72 for evaluation. The largest system contains 77 state variables evaluated at 101 requested time points, resulting in a $101\times77$ trajectory matrix as the solver output.

% The task is to generate a general-purpose numerical integrator for initial-value problems of the form
% \[\dot{\mathbf{y}}(t)=f(t,\mathbf{y}(t);\boldsymbol{\theta}).\]

\vspace{-0.5em}
\paragraph{Training \& evaluation setup.} These reaction-network systems often exhibit stiffness, making reliable time integration and error control essential. Each solver is provided with the same fixed right-hand-side (RHS) implementation and is required to implement the integration algorithm using only Python and NumPy. A case is considered successful only if the solver converges, returns a finite trajectory on the requested time grid, remains within the computational budget, and satisfies
\(
\max_{i,j}\tfrac{|\hat{y}_i(t_j)-y_i^{\text{ref}}(t_j)|}{\max(1,|y_i^{\text{ref}}(t_j)|)}\leq10^{-4}
\) over all reference-reported states and time points. Here, $y_i^{\mathrm{ref}}(t_j)$ denotes the reference trajectory provided by the benchmark. We adopt $10^{-4}$ as the trajectory-error threshold, following the accuracy criterion used in the original benchmark \citep{stadter2021benchmarking}. 

Evaluation results are reported in Table~\ref{tab:ode-main-results}, including pass rate, median normalized per-case error, and wall-clock runtime across all hidden evaluation models. Across all agent baselines, ADSD consistently improves solver performance, producing integrators with lower trajectory error and faster runtime.

\begin{table*}[t]
\centering
\caption{Performance on held-out stiff-ODE cases from 72 source families with 117 models. $N_0$ denotes the solver generated from the agent's inherent knowledge, and ADSD denotes the fresh solver generated with the constructed skill packet. Error is the median trajectory error. Wall time usages are per-case mean $\pm$ standard deviation across all attempts.}
\label{tab:ode-main-results}
\small
\setlength{\tabcolsep}{8pt}
\resizebox{0.9\linewidth}{!}{%
\begin{tabular}{llccc}
\toprule
\textbf{Base Model}
& \textbf{Configuration}
& \textbf{Pass rate} $(\uparrow)$
& \textbf{Median error} $(\downarrow)$
& \textbf{Time} $(\downarrow)$\\
\midrule

\multirow{2}{*}{\texttt{GPT-5.5}}
& $N_0$ solver
& $70.1\% $
& $9.39\times10^{-6}$
& $4.713 \pm 9.570$ \\

& \quad + ADSD
& $90.6\%$
& $3.93\times10^{-7}$
& $3.451 \pm 6.743$ \\

\midrule

\multirow{2}{*}{\texttt{Claude-Sonnet-5}}
& $N_0$ solver
& $94.0\%$
& $4.43\times10^{-7}$
& $0.618 \pm 1.457$ \\

& \quad + ADSD
& $100.0\%$
& $5.04\times10^{-8}$
& $0.496 \pm 0.782$ \\

\midrule

\multirow{2}{*}{\texttt{Qwen3.8-2.4T}}
  & $N_0$ solver
  & $98.3\%$
  & $1.30\times10^{-8}$
  & $0.961 \pm 1.688$ \\

  & \quad + ADSD
  & $99.2\%$
  & $5.95\times10^{-9}$
  & $0.646 \pm 1.083$ \\

\bottomrule
\vspace{-3em}
\end{tabular}%
}
\end{table*}

\vspace{-0.5em}
\subsection{Elliptic PDE solver}\label{sec:pde}

\begin{table*}[t]
  \centering
  \caption{Performance on 2,400 held-out ConDiff cases. Errors are relative to the benchmark's discrete reference solutions. All error, wall-time, and operator-call statistics are mean $\pm$ population std across 7,200 runs per solver (three repetitions per case). }
  \label{tab:pde-main-results}
  \small
  \setlength{\tabcolsep}{6pt}
  \resizebox{\linewidth}{!}{%
  \begin{tabular}{llcccc}
  \toprule
  \textbf{Base Model}
  & \textbf{Configuration}
  & \textbf{$\ell_2$ error} $(\downarrow)$
  & \textbf{$\ell_\infty$ error} $(\downarrow)$
  & \textbf{Time} $(\downarrow)$
  & \textbf{Operator calls} $(\downarrow)$ \\
  \midrule

  \multirow{2}{*}{\texttt{GPT-5.5}}
  & $N_0$ solver
  & $1.69\times10^{-9} \pm 0.86\times10^{-9}$
  & $1.46\times10^{-9} \pm 0.91\times10^{-9}$
  & $7.29 \pm 1.32$
  & $215.49 \pm 40.35$ \\

  & \quad + ADSD
  & $5.27\times10^{-10} \pm 2.04\times10^{-10}$
  & $5.93\times10^{-10} \pm 3.33\times10^{-10}$
  & $2.78 \pm 0.58$
  & $71.22 \pm 17.58$ \\

  \midrule
  \multirow{2}{*}{\texttt{Claude-Sonnet-5}}
  & $N_0$ solver
  & $1.12\times10^{-11}\pm0.54\times10^{-11}$ & $9.96\times10^{-12}\pm6.12\times10^{-12}$& $0.86\pm0.03$ & $148.16\pm4.82$ \\

  & \quad + ADSD
  & $8.15\times10^{-12}\pm5.41\times10^{-12}$ & $5.49\times10^{-12}\pm4.97\times10^{-12}$& $0.79\pm 0.02$& $125.71\pm13.40$ \\

  \midrule
  \multirow{2}{*}{\texttt{Qwen3.8-2.4T}}
  & $N_0$ solver
  & $6.49\times10^{-9}\pm3.53\times10^{-9}$ & $4.52\times10^{-9}\pm2.83\times10^{-9}$& $0.70\pm0.09$ & $86.81\pm 29.68$\\

  & \quad + ADSD &
$3.55\times10^{-10}\pm2.58\times10^{-10}$ & $3.72\times10^{-10}\pm2.79\times10^{-10}$ & $0.64\pm0.02$ & $17.57\pm10.03$ \\
  \bottomrule
  \vspace{-3em}
  \end{tabular}%
  }
  \label{tab:pderesults}
  \vspace{-0.5em}
\end{table*}

We select the heterogeneous diffusion benchmark from ConDiff \citep{trifonov2024condiff}, which asks the agent to generate a reusable numerical solver for the two-dimensional steady-state diffusion problems defined in \cref{eq:pde}. We use the benchmark's $128\!\times\!128$ diffusion subset, comprising 12 coefficient configurations with different spatial correlations and contrast levels. The testcase contains 12,000 training cases and 2,400 held-out evaluation cases, with 200 evaluation cases per configuration. Evaluation therefore measures generalization to unseen coefficient and forcing realizations within the same configuration families. Each case requires solving for 16,384 unknowns on a $128\!\times\!128$ grid.

\vspace{-0.5em}
\paragraph{Training \& evaluation setup.} The heterogeneous, high-contrast coefficients can induce ill-conditioned linear systems, making effective preconditioning and convergence control essential. Each solver receives the coefficient field, forcing vector, and the same fixed implementation of the benchmark's discrete diffusion operator. Denoting the resulting system by $A\mathbf{u}=\mathbf{b}$, a run is considered successful only if the solver converges and returns a finite solution. We evaluate each solver-case pair using three fresh-process repetitions with alternating solver order, and a case passes only if all three repetitions succeed. For solution accuracy, we report both $\ell_2$ and
$\ell_\infty$ errors:
\(
  e_p =\tfrac{\|\hat{\mathbf{u}}-\mathbf{u}^{\mathrm{ref}}\|_p}
       {\|\mathbf{u}^{\mathrm{ref}}\|_p}\), \(p\in\{2,\infty\},
\)
where $\mathbf{u}^{\mathrm{ref}}$ is the discrete reference solution provided by ConDiff, and $e_p$ quantify the overall solution error and the largest absolute component error normalized by the reference solution's maximum magnitude, respectively. Results are summarized in Table~\ref{tab:pderesults}. Among all baselines, ADSD effectively improves the solver performance.

\subsection{Additional Analyses} \label{sec:aa}

We conduct two additional analyses to assess whether ADSD provides benefits beyond solver revision using only a model's inherent knowledge.

\vspace{-0.5em}
\paragraph{Ablation analysis on ADSD.} We first isolate the contribution of ADSD's discovered skills by removing the ADSD pipeline and directly asking each agent to improve its frozen baseline solver $N_0$, without providing any diagnosed hypotheses or discovered skills. All remaining prompt instructions, model configurations, reasoning budgets, and runtime constraints are kept unchanged. We evaluate this ablation in two numerical domains: PF$\Delta$ from power-grid engineering and stiff-ODE integration from dynamical-system engineering.

\begin{figure*}[t]
    \centering
\includegraphics[width=\linewidth]{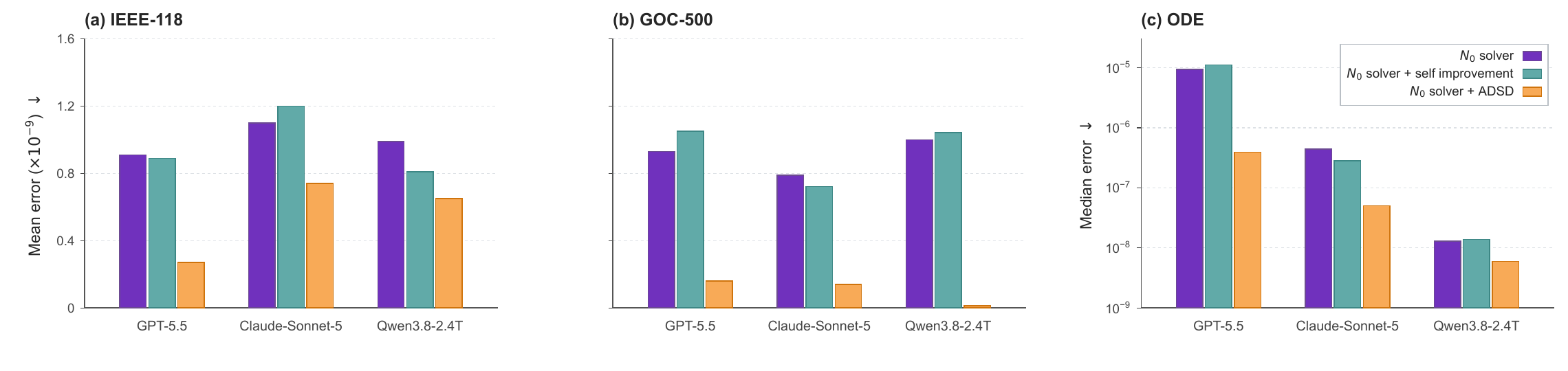}
\vspace{-2.5em}
    \caption{Comparison of the $N_0$ solver, self improvement, and ADSD on (a) IEEE-118, (b) GOC-500, and (c) stiff ODEs. Bars show mean error for grid tasks and median trajectory error for ODEs. Panel (c) uses a logarithmic scale; lower is better throughout.}
    \label{f3}
    \vspace{-0.5em}
\end{figure*}

Figure~\ref{f3} compares the original $N_0$ solver, direct self-improvement, and ADSD across three base models. Direct self-improvement produces inconsistent gains, which error decreases in four comparisons but increases in the remaining five. Still, ADSD achieves significantly lower error in all settings. These results indicate that simply requesting another solver revision does not reliably improve numerical performance, whereas the diagnostic hypotheses and literature-grounded skills discovered by ADSD provide substantially more consistent improvements across the evaluated models and domains.

\vspace{-0.5em}
\paragraph{ADSD versus recursive self-improvement.} The previous ablation considers only one round of direct self-improvement, raising the question of whether repeated revisions could eventually achieve comparable performance. We therefore extend the same procedure to five iterations, where at each iteration the agent can receive the solvers from all previous rounds.

\begin{figure*}[t]
\centering
\includegraphics[width=0.99\linewidth]{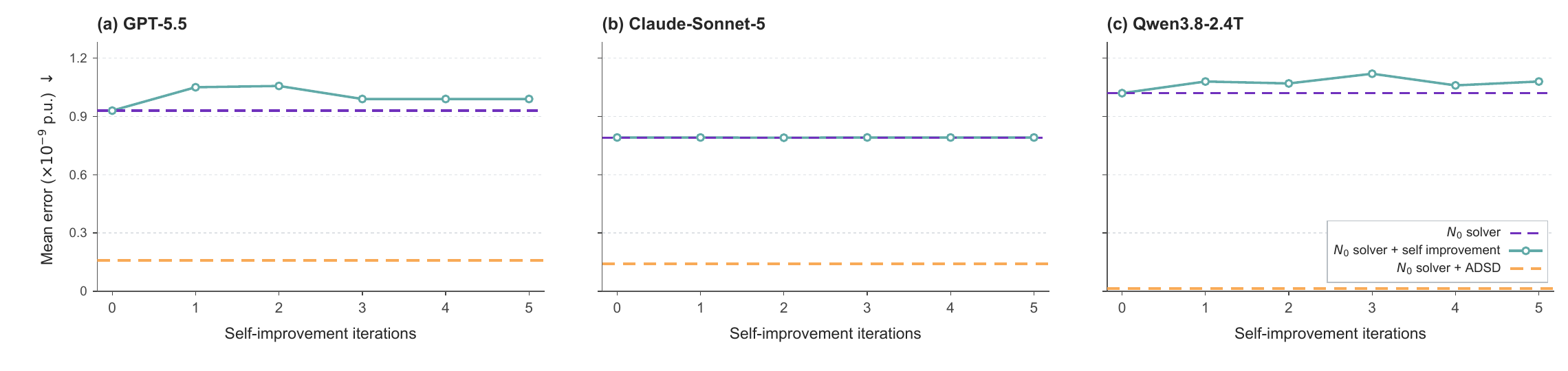}
\vspace{-1.5em}
\caption{Mean error on GOC-500 across recursive self-improvement iterations for (a) GPT-5.5, (b) Claude-Sonnet-5, and (c) Qwen3.8-2.4T. Dashed lines indicate the $N_0$ and ADSD baselines. Iteration 0 denotes the original solver; lower is better.}
\label{f4}
\vspace{-1.5em}
\end{figure*}

Shown in Figure~\ref{f4}, extending direct self-improvement to more iterations does not produce sustained error reduction. GPT baseline initially worsens, then partially recovers before plateauing above the error of its original $N_0$ solver, while Claude-Sonnet-5 remains nearly unchanged across iterations. Thus, within the evaluated iteration budget, additional rounds of self-diagnosis and solver revision using inherent model knowledge alone do not close the performance gap to ADSD. Together with the single-round ablation, these results support the contribution of ADSD's externally discovered, diagnosis-targeted skills beyond simply allocating additional opportunities for solver revision.

\section{Conclusion}

In Appendix~\ref{app:exp}, we provide more discussions and backgrounds for readers, along with a further analysis of the necessity for agent to do diagnosis, which highlights that access to numerical knowledge alone does not determine which method addresses a specific solver failure. 

We introduced ADSD, an automatic framework that turns observed solver behavior into verified numerical conditions and reusable executable skills. ADSD grounds solver improvement in literature-derived diagnostic probes, hypothesis verification, and final skill construction. Across four challenging numerical domains, ADSD improves solver accuracy, robustness, and efficiency across different base models, with learned skills transferring to unseen problem regimes. Our ablations show that direct or recursive self-improvement using the model’s inherent knowledge does not recover these gains within the evaluated budgets. Together, these results highlight diagnosis as a critical component of numerical intelligence, connecting observed solver behavior to actionable numerical knowledge. ADSD offers a practical path toward scientific agents that identify why a solver struggles, acquire appropriate methods, and retain them as reusable skills for improving the underlying numerical algorithms.

\newpage
\bibliography{iclr2026_conference}
\bibliographystyle{iclr2026_conference}

\newpage
\appendix
\section{Further Related Works}\label{app:rw}

\vspace{-0.5em}
\paragraph{Power-system knowledge and applications.}
Foundation-model applications, domain-adapted models such as GAIA, and dispatch benchmarks, such as ElecBench, explore power-system knowledge and reasoning \citep{huang2023large,cheng2024gaia,zhou2024elecbench}. Related studies enable LLMs to operate previously unseen simulation tools through documentation and prompting \citep{jia2024enabling}, or interpret observations for grid-fault diagnosis \citep{jing2024fault}. These tasks concern domain understanding, tool use, and physical-system conditions, while our work instead focuses on numerical conditions associated with solver behavior.

\vspace{-0.5em}
\paragraph{Model discovery and formulation.}
An upstream problem is determining the mathematical model to be solved. SINDy identifies governing equations through sparse regression, while LLM-DMD uses language models to guide power-system dynamic model discovery \citep{brunton2016sindy,shen2026llmdmd}. Other approaches translate natural-language requirements into optimization formulations and validate the generated models \citep{hu2025natural,shen2026proopf}. Mamo \citep{huang2025mamo} similarly evaluates mathematical modeling for ODEs and optimization. For power models specifically, L2RPN \citep{marot2021learning} and RL2Grid \citep{marchesini2025rl2grid} study reinforcement learning for sequential grid-operation decisions. \citet{fan2022powergym} and \citet{biagioni2022powergridworld} also provides the gym for training power agents.

\vspace{-0.5em}
\paragraph{Numerical methods and solver evaluation.} Classical numerical analysis provides the methods that solver-building agents can acquire. Newton power flow and higher-order methods with factorization reuse form relevant foundations \citep{tinney1967power}. Automated integrator design also predates LLMs: \citet{guo2022personalized} learn Runge--Kutta coefficients for targeted ODE families. For evaluation, \citet{stadter2021benchmarking} compare integration methods and configurations on biological ODE models, while PETSc agent evaluation and PDEAgent-Bench assess generated scientific code for numerical accuracy, efficiency, and appropriate use of professional libraries \citep{zhang2026agentic,hang2026pdeagent}. We note that these studies emphasize that successful execution alone does not establish numerical quality.

\vspace{-0.5em}
\paragraph{Scientific retrieval.} We further discuss ADSD's scope to scientific retrieval literature. Recently, MIR \citep{garikaparthi2025mir} retrieves methodological inspirations, and OpenScholar \citep{asai2024openscholar} as well as PaperQA2 \citep{skarlinski2024language} support literature-grounded scientific synthesis. The AI Scientist \citep{lu2024ai} and Agent Laboratory \citep{schmidgall2025agent} extend automation across literature review, experimentation, and research reporting. FlamePilot \citep{xiao2026towards} also connects literature with self-corrective combustion simulation workflows. For ADSD, it specializes this connection between literature and computation by checking a numerical method's assumptions against an existing solver before constructing a transferable implementation.

We note that ADSD is complementary to these general skill-discovery frameworks. Paper2Agent \citep{miao2025paper2agent} converts a specified paper and its codebase into executable tools, SkillFoundry \citep{shen2026skillfoundry} mines and validates skills from heterogeneous scientific resources, and Voyager \citep{wang2025voyager} acquires reusable skills through execution feedback. Our skill-discovery stage (\S\ref{sec:SD}) shares their goal of constructing reusable executable knowledge. ADSD's distinguishing focus is the preceding diagnostic step: it converts a frozen solver's observed behavior into verified numerical hypotheses that specify which methods to retrieve and which assumptions their implementations must satisfy. This provides a concrete connection between solver failures and skill acquisition, making existing skill-construction mechanisms complementary to the diagnosis front-end.

\vspace{-0.5em}
\paragraph{Agentic skills aligned from experience.} Finally, we go over prior classic works to incentive agent's skill and ability. Reflexion \citep{shinn2023reflexion} and ExpeL \citep{zhao2024expel} first turn execution feedback and accumulated experience into reusable guidance for subsequent tasks. Large Language Models as tool makers separates executable-tool construction from downstream tool use \citep{cai2024large}. Further works also studies reinforcement learning to improve the agent's policy through skill, experience, and learning from mistakes \citep{wang2026reinforcement, chen2026stepwise}, which establishes fundamental routes for retaining knowledge through memory, executable tools, and environment interactions \citep{guo2025genenv}. Apart from these in-context agentic tasks, recent LLM-RL framework also introduces also introduces language model to further incentive language model's ability for mathematics \citep{Li-2024-Remax,Yu-2025-Dapo,Chen-2025-Exploration,Shao-2024-Deepseekmath} and skill-based tool usage \citep{qian2026toolrl}; We believe it is a promising direction to extend ADSD to these domains as a self-supervised pipeline as a replacement the GPU-extensive online RL training. Lastly, language alignment methods \citep{Ouyang-2022-Training,Rafailov-2023-Direct,Meng-2024-SimPO,Chen-2025-ComPO,chen2026reward} also incurs online RL training, and how to adapt ADSD to the structure with contrastive signals and align language model would also be a valuable direction to study.

\section{Technical case studies}\label{app:case}

\subsection{GPT baseline over PF$\Delta$}\label{app:gpt-pf}
We provide a concrete walkthrough of the self-evolution process from \S\ref{sec:exp-pf} using the GPT baseline on the IEEE 118-bus training environment, which includes the initial $N_0$ solver, the diagnosis skills learned, the initial symptoms identified together with the final verified hypotheses, the resulting skills built, and the final improved solver.

\vspace{-0.5em}
\paragraph{Initial $N_0$ solver.} Before presenting the subsequent diagnosis and solver improvement, we first describe the baseline implementation produced by the agent. The resulting implementation is a dense polar Newton power-flow solver with an analytic Jacobian, a safeguarded step, and a bounded regularized fallback for singular linear systems.

Here is the core implementation of the $N_0$ solver:

\begin{lstlisting}[
language=Python,
escapeinside={(@}{@)}
]
for _ in range(MAX_ITERS):
    j  = jacobian(parsed, va, vm)                 (@\hfill\textcolor{red}{\# one assembly}@)
    dx = solve_linear_system(j, mismatch, diag)   (@\hfill\textcolor{red}{\# one factorization}@)
    if dx is None:
        return failed_result(case_id, "jacobian solve failed", diag)

    base_scale = trust_region_scale(parsed, dx)   (@\hfill\textcolor{red}{\# $\leq 0.5$ rad, $\leq 0.2$ p.u.}@)
    accepted   = line_search(parsed, va, vm, norm, dx, base_scale)
    if accepted is None:
        return failed_result(case_id, "line search failed", diag)

    va, vm, mismatch, norm, scale, step_norm = accepted
    diag["nonlinear_iterations"] += 1
    if norm <= POWER_TOL:                         
        return success_result(parsed, va, vm, diag)
\end{lstlisting}

Every accepted nonlinear iteration of $N_0$ costs exactly one Jacobian assembly, one dense factorization, and one triangular solve. Therefore, for $N_0$, ``one more Newton iteration'' and ``one more dense factorization'' are the same statement. This is what makes the diagnosis we next introduced actionable, and it is exactly the quantity that the two solver skills attack. Apart from the solver itself, $N_0$ also records the following execution trace that the diagnosis stage utilizes for analysis: per-case
counters for nonlinear iterations, Jacobian evaluations, factorizations, linear solves, fallback activations, and factorization failures, together with residual, step-norm, and accepted-step-scale histories.

\vspace{-0.5em}
\paragraph{Diagnosis skills \& hypothesis generation.} Following Algorithm~\ref{alg:auto-diagnosis}, the agent constructs ten deterministic diagnosis skills from the literature, each mapping a sanitized testcase. One of them, \texttt{ac-jacobian-criticality}, is especially important, which converts the qualitative notion of a ``weak Jacobian'' into a deterministic feature evaluated on the state that $N_0$ actually returned,
\begin{equation}
\sigma_{\min}\!\left(
J_{\text{calc}}(x_i^{N_0};\xi_i)
\right),
\label{eq:sigma-min}
\end{equation}
and the agent then proposes the following detector
\begin{equation}
\underbrace{
\sigma_{\min}\!\left(
J_{\text{calc}}(x_i^{N_0};\xi_i)
\right)
\leq 0.16
}_{\text{condition }C}
\quad\Longrightarrow\quad
\underbrace{\text{$N_0$ requires more nonlinear iterations}}_{\text{behavior }B}.
\label{eq:detector}
\end{equation}
By examining the agent's action history, we find that the threshold $0.16$ was inferred during hypothesis discovery rather than taken from the literature or hard-coded into the diagnostic skill. Specifically, during the 200-case diagnosis round, the agent observes a lower-quartile Jacobian minimum singular value of approximately $0.16008$ and proposes the nearby cutoff $0.16$. This cutoff identifies 50 cases requiring $5.78$ nonlinear iterations on average, compared with $5.01$ iterations for the remaining cases, providing initial evidence for the hypothesized correlation. The agent then tests nearby cutoffs of $0.15$ and $0.17$ to assess the robustness of this relationship, before further confirming that the same correlation holds on the subsequent 4,000-case verification pool. After these checks, the $0.16$ threshold is frozen for subsequent verification, yielding the final verified hypothesis.

This directly sets the literature search question: which reproducible methods achieve more nonlinear progress per Jacobian factorization while preserving the frozen $N_0$ equations along with the error reduction target? The search screens $312$ clean papers down to $82$ candidates and six fully read selections, producing $40$ verified claims and nine method reviews, from which two methods proceed to executable solver-skill packets: \texttt{multistep-factorization-reuse}, which shares one factorization across several corrections, and \texttt{seventh-order-method9}, which uses two reusable factorizations in a higher-order
update.

\begin{table*}[t]
\centering
\caption{Solver-skill ablation on IEEE-118 validation set. Each $B$ variant is separately generated with mandatory integration of the indicated skills. Factorizations, linear solves, runtime, and residual are per-case means. The residual is the independently evaluated $\|F_{\mathrm{PF}}(x;\xi)\|_{\infty}$.}
\label{tab:ablation}
\small
\setlength{\tabcolsep}{8pt}
\resizebox{\linewidth}{!}{
\begin{tabular}{lccrrrr}
  \toprule
  Solver
    & \texttt{Multistep}
    & \texttt{Method 9}
    & Factorizations
    & Linear solves
    & Runtime (s)
    & PF residual \\
  \midrule
  $N_0$
    & -- & --
    & 5.116
    & 5.116
    & 0.727
    & $9.10\times10^{-10}$ \\
  $B_1$
    & $\checkmark$ & --
    & 2.755
    & 8.255
    & 0.494
    & $9.86\times10^{-10}$ \\
  $B_2$
    & -- & $\checkmark$
    & 4.003
    & 8.006
    & 0.654
    & $7.46\times10^{-13}$ \\
  $B_3$
    & $\checkmark$ & $\checkmark$
    & 3.114
    & 7.114
    & 0.538
    & $2.76\times10^{-10}$ \\
  \bottomrule
  \vspace{-3em}
\end{tabular}
}
\end{table*}

\vspace{-0.5em}
\paragraph{Solver skills.} With two generated skill packages, the agent made three changes to the $N_0$ solver: (i) the correction rule applied inside a single outer iteration; (ii) the baseline's Gaussian elimination, which is separated into a reusable factorization followed by forward and backward substitution; (iii) and the trust region and backtracking line search, which on the fast path are replaced by an explicit accept-reject test. The ablation for the two skills are shown in Table~\ref{tab:ablation}, where we can clearly tell that skill \texttt{Multistep} is specifically for improving the solver efficiency while maintaining the error performance, and skill \texttt{Method 9} is focusing more on improving the solver performance.

We first describe the first skill: \textbf{Multistep Factorization Reuse}, which targets the repeated factorization cost of $N_0$ directly. Instead of factorizing a new Jacobian for every correction, it fixes the Jacobian at the current outer iterate and reuses its factors across several internal corrections. The evaluated implementation uses four stages. Starting with $y_{k,1}=x_k$, it computes
\begin{equation}
\begin{aligned}
m_{k,i}
&= \frac{1}{i-1}
\sum_{j=1}^{i-1} y_{k,j},
&& i=2,3,4, \\
J(x_k)\,d_{k,i}
&= F(m_{k,i}),
&& i=2,3,4, \\
y_{k,i}
&= m_{k,i}+i\,d_{k,i},
&& i=2,3,4, \\
\widehat{x}_{k+1}
&= \frac{1}{4}\sum_{j=1}^{4}y_{k,j}.
\end{aligned}
\label{eq:multistep}
\end{equation}
The positive correction sign follows the specified-minus-calculated mismatch convention of $N_0$. All three linear systems share the same matrix $J(x_k)$, so one outer update now requires one Jacobian assembly, one factorization, and three right-hand-side solves. By tracing the agent thinking traces, we find this skill is learnt by reading \citet{tostado2020novel} from the literature space; specifically, the agent extracted the multistep update from Section II-B, Eqs.~(4a)--(4c).

We next introduce the second skill: \textbf{Seventh-order Method 9}, which instead buys a higher convergence order using two reusable Jacobian factorizations. Its four corrections are
\begin{equation}
\begin{aligned}
y_k
&= x_k + J(x_k)^{-1}F(x_k), \\
z_k
&= y_k + J(x_k)^{-1}F(y_k), \\
m_k
&= \frac{y_k+z_k}{2}, \\
w_k
&= y_k + J(m_k)^{-1}F(y_k), \\
\widehat{x}_{k+1}
&= w_k + J(m_k)^{-1}F(w_k).
\end{aligned}
\label{eq:method9}
\end{equation}
Here, inverse notation denotes linear solves, as the implementation does not explicitly form matrix inverses. The first factorization, at $x_k$, is reused to compute $y_k$ and $z_k$; the second, at the midpoint $m_k$, is reused to compute $w_k$ and $\widehat{x}_{k+1}$. One outer update therefore requires two assemblies, two factorizations, and four linear solves, using residuals at $x_k$, $y_k$, and $w_k$, followed by an additional residual evaluation to check the proposed outer state. By tracing the agent's trajectories, we find that this skill is learnt from \citet{tostado2020two}, where the agent extracted Eq.~(9) and Algorithm 1 in Section IV-B. 

\subsection{Qwen baseline over AC-OPF}\label{app:qwen}

We walk through the second case study, with nearly $84\%$ pass rate improvement over the $N_0$ solver. This time, the domain is constrained optimization rather than the power-equation solving of \S\ref{app:gpt-pf}, and the baseline is Qwen. Recall that in \S\ref{app:gpt-pf}, the agent learns \emph{cost} strategies: $N_0$ already reaches the solution, and the skills make each step toward it cheaper and more accurate. Here instead, it learns a \emph{scaling} strategy that genuinely improve the qualitative behavior of the solver.

\vspace{-0.5em}
\paragraph{Initial $N_0$ solver.} The agent's baseline is a dense primal-dual
interior-point method with six main steps: explicit inequality slacks, analytic first and second derivatives, a fraction-to-boundary line search, a regularized fallback for the KKT system, retention of the best iterate seen, and a stall exit. Its outer loop is governed by a scalar first-order progress measure:

\begin{lstlisting}[
language=Python,
escapeinside={(@}{@)}
]
scale = 1.0 + float(np.max(np.abs(gf)))       (@\hfill\textcolor{red}{\# $1+\|\nabla f\|_\infty$}@)
for it in range(max_iter):                    (@\hfill\textcolor{red}{\# max\_iter $=200$}@)
    ...
    meas = max(feas_eq, feas_ineq,            (@\hfill\textcolor{red}{\# scalar progress measure}@)
               statn / scale, comp_max)
    if meas < best_meas:                      (@\hfill\textcolor{red}{\# retain best iterate}@)
        best_meas, best_state = meas, (x.copy(), y.copy(), ...)
    if converged:
        break
    if meas < best_meas_prev * (1.0 - 1e-3):  (@\hfill\textcolor{red}{\# $10^{-3}$ relative progress}@)
        stalled = 0
    else:
        stalled += 1
    best_meas_prev = best_meas
    if stalled > 12:                          (@\hfill\textcolor{red}{\# patience $=12$}@)
        result.reason = "stalled"
        break
...
if result.status != "converged" and best_state is not None:
    x, y, pi, zL, zU, sigma = best_state[:6]  (@\hfill\textcolor{red}{\# return the best iterate}@)
\end{lstlisting}

We note that two properties of this loop matter for what follows: (i) $\texttt{meas}$ aggregates four quantities of different units into one scalar, and normalizes the stationary residual by $1+\|\nabla f\|_\infty$, evaluated at the current iterate; (ii) termination is decided by relative improvement of the running best of that scalar, with a patience of $12$. 

\vspace{-0.5em}
\paragraph{Diagnosis skills \& hypothesis generation.} We just directly jump into the core part of the diagnosis result, and how it is generated follows from the same procedure in \S\ref{app:gpt-pf}. The most important hypothesis, $H_1$, is built from a distributional invariance that the agent reads directly off the trace. On the failing rows the reported stationarity residual lies in $[3718.78,\,3718.943]$, within $3.17\times10^{-5}$ relative of
\begin{equation}
\|\nabla f(x_0)\|_\infty \;=\; 3718.8979,
\label{eq:invariant}
\end{equation}
which is the same constant on every case in the population, because the IEEE-57 generator cost vector is linear and invariant to both the contingency and the load draw. A residual pinned at the raw cost gradient is the signature of a point at which the multipliers are still essentially zero. So, the agent therefore proposes the following symptom-condition hypothesis:
\begin{equation}
\underbrace{
\#\{\texttt{ipm.py::\_solve\_kkt}\}\;\leq\;27
}_{\text{condition }C}
\quad\Longrightarrow\quad
\underbrace{\text{$N_0$ does not converge}}_{\text{behavior }B},
\label{eq:detector-opf}
\end{equation}
If we get back to the $N_0$ solver, from the
traced first-iteration state: initial dual steps of magnitude
$1.81\times10^{6}$--$2.17\times10^{7}$ drive the fraction-to-boundary rule to a primal step fraction of $2.67\times10^{-8}$--$1.50\times10^{-7}$, so the iterate does not move, and $\texttt{meas}$ never improves by the required $10^{-3}$ for $13$ consecutive iterations. On the discovery split, the condition selects $855$ rows at a nonconvergence rate of $1.000$ against $0.0276$ for the $145$ controls; on the held-out verification split, it selects $833$ rows, of which $832$ fail, against $4$ of $167$ controls, a difference of $+97.48$ percentage points with $95\%$ interval $[95.15,\,99.82]$. Therefore, this motivates agent to propose the literature search question: which published mechanism makes an interior-point solver's first-order quantities commensurate, and which normalization keeps a first-order residual interpretable when the objective gradient norm is population-invariant?

\vspace{-0.5em}
\paragraph{Solver skill generation.} With the search question proposed, the agent screens $376$ papers across $11$ directions with $4{,}136$ screening judgments, retaining $49$ direct and $156$ possible candidates, and places $43$ primaries over $20$ unique papers. The final skill located is \texttt{gmax-rowscale-kkt-measure}, which is learned from an old but very effective paper \citep{wachter2006implementation}. The agent combines two elements of that paper which $N_0$ uses neither of. The first is the one-time automatic row scaling in \S3.8 of \citet{wachter2006implementation}.  With $g_{\max}=100$ and $D_x=I$, computed once at the sanitized start, it has
\begin{equation}\label{eq:rowscale}
d_f=\min\!\Big\{1,\frac{g_{\max}}{\|\nabla f(x_0)\|_\infty}\Big\},\ 
d_e^{(j)}=\min\!\Big\{1,\frac{g_{\max}}{\|A_e(x_0)_{j,:}\|_\infty}\Big\},\
d_i^{(j)}=\min\!\Big\{1,\frac{g_{\max}}{\|A_i(x_0)_{j,:}\|_\infty}\Big\},
\end{equation}
which is applied as $\tilde f=d_f f$, $\tilde c_e=D_e c_e$, $\tilde c_i=D_i c_i$ with the variable bounds untouched. Because the factors are positive and diagonal this is an equivalence transformation: same feasible set, same argmin, and multipliers and the Lagrangian gradient map back exactly by $y=D_e\tilde y/d_f$, $\pi=D_i\tilde\pi/d_f$, $r_{\text{stat}}=\tilde r_{\text{stat}}/d_f$. On this population, the rule is almost entirely a statement about the objective: the $434$ constraint rows have median $\infty$-norm exactly $1.00$ and maximum $138.27$, so only $2$ of them are rescaled at all, while $\|\nabla f(x_0)\|_\infty$ is the constant of \cref{eq:invariant}, which is roughly $3{,}700\times$ the typical constraint row and $27\times$ the largest. The rule therefore proceeds as 
\begin{equation}
d_f=\frac{100}{3718.8979}=2.68897\times10^{-2},
\end{equation}
dividing the objective row by $37.2$ and leaving $432$ of $434$ constraint rows untouched.

\begin{table*}[t]{
\vspace{-1em}
\centering
\caption{
Topology generalization from the feasible \texttt{N2}-topology grid to previously unseen feasible \texttt{N1} and full-grid \texttt{N} cases. Values are mean \(\pm\) std over 10,000 cases per generalization topology.
}
\label{tab:topology-generalization}
\small
\setlength{\tabcolsep}{4.5pt}
\resizebox{\linewidth}{!}{
\begin{tabular}{lll|cc|cccc}
\toprule
\multirow{2}{*}{\textbf{Base Model}}
& \multirow{2}{*}{\textbf{Dataset}}
& \multirow{2}{*}{\textbf{Configuration}}
& \multicolumn{2}{c}{\textbf{In-domain \texttt{N2}}}
& \multicolumn{2}{c}{\textbf{Generalization \texttt{N1}}}
& \multicolumn{2}{c}{\textbf{Generalization \texttt{N}}} \\
\cmidrule(lr){4-5}\cmidrule(lr){6-7}\cmidrule(lr){8-9}
&&&
\textbf{Time} \((\downarrow)\) &
\textbf{Error} \((\downarrow)\) &
\textbf{Time} \((\downarrow)\) &
\textbf{Error} \((\downarrow)\) &
\textbf{Time} \((\downarrow)\) &
\textbf{Error} \((\downarrow)\) \\
\midrule
\multirow{4}{*}{\texttt{GPT-5.5}}
& \multirow{2}{*}{IEEE-118}
& \(N_0\) solver
& \(0.73 \pm 0.21\)
& \(9.1{\times}10^{-10} \pm 2.0{\times}10^{-9}\)
& \(0.74 \pm 0.21\)
& \(7.6{\times}10^{-10} \pm 1.8{\times}10^{-9}\)
& \(0.75 \pm 0.20\)
& \(8.9{\times}10^{-10} \pm 2.0{\times}10^{-9}\) \\
&& \quad + ADSD
& \(0.55 \pm 0.15\)
& \(2.7{\times}10^{-10} \pm 9.9{\times}10^{-10}\)
& \(0.54 \pm 0.15\)
& \(2.1{\times}10^{-10} \pm 9.1{\times}10^{-10}\)
& \(0.53 \pm 0.14\)
& \(2.3{\times}10^{-10} \pm 8.3{\times}10^{-10}\) \\
\cmidrule(lr){2-9}
& \multirow{2}{*}{GOC-500}
& \(N_0\) solver
& \(0.51 \pm 0.14\)
& \(9.3{\times}10^{-10} \pm 2.2{\times}10^{-9}\)
& \(0.38 \pm 0.09\)
& \(1.2{\times}10^{-9} \pm 2.5{\times}10^{-9}\)
& \(0.37 \pm 0.09\)
& \(8.6{\times}10^{-10} \pm 2.1{\times}10^{-9}\) \\
&& \quad + ADSD
& \(0.50 \pm 0.14\)
& \(1.6{\times}10^{-10} \pm 7.4{\times}10^{-10}\)
& \(0.62 \pm 0.15\)
& \(1.1{\times}10^{-10} \pm 5.4{\times}10^{-10}\)
& \(0.60 \pm 0.15\)
& \(1.9{\times}10^{-10} \pm 7.9{\times}10^{-10}\) \\
\midrule
\multirow{4}{*}{\texttt{Claude-Sonnet-5}}
& \multirow{2}{*}{IEEE-118}
& \(N_0\) solver
& \(0.16 \pm 0.04\)
& \(1.1{\times}10^{-9} \pm 2.1{\times}10^{-9}\)
& \(0.18 \pm 0.04\)
& \(1.1{\times}10^{-9} \pm 1.9{\times}10^{-9}\)
& \(0.13 \pm 0.01\)
& \(1.1{\times}10^{-9} \pm 1.9{\times}10^{-9}\) \\
&& \quad + ADSD
& \(0.15 \pm 0.04\)
& \(7.4{\times}10^{-10} \pm 1.8{\times}10^{-9}\)
& \(0.17 \pm 0.04\)
& \(7.2{\times}10^{-10} \pm 1.8{\times}10^{-9}\)
& \(0.12 \pm 0.01\)
& \(8.8{\times}10^{-10} \pm 1.9{\times}10^{-9}\) \\
\cmidrule(lr){2-9}
& \multirow{2}{*}{GOC-500}
& \(N_0\) solver
& \(0.52 \pm 0.26\)
& \(7.9{\times}10^{-10} \pm 1.9{\times}10^{-9}\)
& \(0.52 \pm 0.16\)
& \(5.1{\times}10^{-10} \pm 1.5{\times}10^{-9}\)
& \(0.54 \pm 0.16\)
& \(7.8{\times}10^{-10} \pm 1.8{\times}10^{-9}\) \\
&& \quad + ADSD
& \(0.52 \pm 0.25\)
& \(1.4{\times}10^{-10} \pm 8.4{\times}10^{-10}\)
& \(0.52 \pm 0.14\)
& \(1.6{\times}10^{-10} \pm 9.1{\times}10^{-10}\)
& \(0.54 \pm 0.15\)
& \(1.7{\times}10^{-10} \pm 8.7{\times}10^{-10}\) \\
\midrule
\multirow{4}{*}{\texttt{Qwen3.8-2.4T}}
& \multirow{2}{*}{IEEE-118}
& \(N_0\) solver
& $0.57 \pm 0.02$
& $9.9{\times}10^{-10} \pm 2.1{\times}10^{-9}$
& \(0.49 \pm 0.01\)
& \(9.5{\times}10^{-10} \pm 2.0{\times}10^{-9}\)
& \(1.21 \pm 0.37\)
& \(9.4{\times}10^{-10} \pm 2.0{\times}10^{-9}\) \\
&& \quad + ADSD
& $0.54\pm 0.01$
& $6.5{\times}10^{-10} \pm 1.6{\times}10^{-9}$ 
& \(0.49 \pm 0.01\)
& \(7.0{\times}10^{-10} \pm 1.7{\times}10^{-9}\)
& \(1.22 \pm 0.35\)
& \(6.1{\times}10^{-10} \pm 1.8{\times}10^{-9}\) \\
\cmidrule(lr){2-9}
& \multirow{2}{*}{GOC-500}
& \(N_0\) solver
& $0.18\pm0.09$  
& $1.0{\times}10^{-9}\pm2.1{\times}10^{-9}$
& \(0.19 \pm 0.09\)
& \(8.2{\times}10^{-10} \pm 1.8{\times}10^{-9}\)
& \(0.20 \pm 0.08\)
& \(8.5{\times}10^{-10} \pm 1.9{\times}10^{-9}\) \\
&& \quad + ADSD
& $0.18\pm0.10$  &  $1.4{\times}10^{-11}\pm2.4{\times}10^{-10}$
& \(0.19 \pm 0.09\)
& \(1.5{\times}10^{-11} \pm 2.5{\times}10^{-10}\)
& \(0.19 \pm 0.09\)
& \(3.7{\times}10^{-11} \pm 3.8{\times}10^{-10}\) \\
\bottomrule
\end{tabular}}
\vspace{-1.5em}
}
\end{table*}

\begin{table*}[t]{
\centering
\caption{
Feasibility generalization from the feasible training set to previously unseen close-to-infeasible grids. Values are mean \(\pm\) std over 5,000 sampled from close-to-infeasible testcases.
}
\label{tab:feasibility-generalization}
\small
\setlength{\tabcolsep}{4.5pt}
\resizebox{\linewidth}{!}{
\begin{tabular}{lll|cc|cc}
\toprule
\multirow{2}{*}{\textbf{Base Model}}
& \multirow{2}{*}{\textbf{Dataset}}
& \multirow{2}{*}{\textbf{Configuration}}
& \multicolumn{2}{c}{\textbf{In-domain \texttt{F}}}
& \multicolumn{2}{c}{\textbf{Generalization \texttt{IF}}} \\
\cmidrule(lr){4-5}\cmidrule(lr){6-7}
&&&
\textbf{Time} \((\downarrow)\) &
\textbf{Error} \((\downarrow)\) &
\textbf{Time} \((\downarrow)\) &
\textbf{Error} \((\downarrow)\) \\
\midrule

\multirow{4}{*}{\texttt{GPT-5.5}}
& \multirow{2}{*}{IEEE-118}
& \(N_0\) solver
& \(0.73 \pm 0.21\)
& \(9.1{\times}10^{-10} \pm 2.0{\times}10^{-9}\)
& \(1.68 \pm 0.24\)
& \(1.1{\times}10^{-9} \pm 2.1{\times}10^{-9}\) \\

&& \quad + ADSD
& \(0.55 \pm 0.15\)
& \(2.7{\times}10^{-10} \pm 9.9{\times}10^{-10}\)
& \(1.25 \pm 0.21\)
& \(3.3{\times}10^{-10} \pm 1.3{\times}10^{-9}\)  \\

\cmidrule(lr){2-7}

& \multirow{2}{*}{GOC-500}
& \(N_0\) solver
& \(0.51 \pm 0.14\)
& \(9.3{\times}10^{-10} \pm 2.2{\times}10^{-9}\)
& \(0.52 \pm 0.14\)
& \(1.1{\times}10^{-9} \pm 2.0{\times}10^{-9}\) \\

&& \quad + ADSD
& \(0.50 \pm 0.14\)
& \(1.6{\times}10^{-10} \pm 7.4{\times}10^{-10}\)
& \(0.53 \pm 0.14\)
& \(8.5{\times}10^{-10} \pm 1.9{\times}10^{-9}\) \\

\midrule

\multirow{4}{*}{\texttt{Claude-Sonnet-5}}
& \multirow{2}{*}{IEEE-118}
& \(N_0\) solver
& \(0.16 \pm 0.04\)
& \(1.1{\times}10^{-9} \pm 2.1{\times}10^{-9}\)
& \(0.73 \pm 0.11\)
& \(1.2{\times}10^{-9} \pm 2.2{\times}10^{-9}\) \\

&& \quad + ADSD
& \(0.15 \pm 0.04\)
& \(7.4{\times}10^{-10} \pm 1.8{\times}10^{-9}\)
& \(0.76 \pm 0.09\)
& \(8.9{\times}10^{-10} \pm 1.6{\times}10^{-9}\) \\

\cmidrule(lr){2-7}

& \multirow{2}{*}{GOC-500}
& \(N_0\) solver
& \(0.52 \pm 0.26\)
& \(7.9{\times}10^{-10} \pm 1.9{\times}10^{-9}\)
& \(1.52 \pm 0.27\)
& \(1.0{\times}10^{-9} \pm 2.0{\times}10^{-9}\) \\

&& \quad + ADSD
& \(0.52 \pm 0.25\)
& \(1.4{\times}10^{-10} \pm 8.4{\times}10^{-10}\)
& \(1.45 \pm 0.26\)
& \(8.1{\times}10^{-10} \pm 1.9{\times}10^{-9}\) \\
\midrule

\multirow{4}{*}{\texttt{Qwen3.8-2.4T}}
& \multirow{2}{*}{IEEE-118}
& \(N_0\) solver
& \(0.57 \pm 0.02\)
& \(9.9{\times}10^{-10} \pm 2.1{\times}10^{-9}\)
& \(1.02 \pm 0.07\)
& \(1.2{\times}10^{-9} \pm 2.1{\times}10^{-9}\) 
\\

&& \quad + ADSD
& \(0.54 \pm 0.01\)
& \(6.5{\times}10^{-10} \pm 1.6{\times}10^{-9}\)
& \(0.92 \pm 0.08\)
& \(7.6{\times}10^{-10} \pm 2.1{\times}10^{-9}\) \\

\cmidrule(lr){2-7}

& \multirow{2}{*}{GOC-500}
& \(N_0\) solver
& \(0.18 \pm 0.09\)
& \(1.0{\times}10^{-9} \pm 2.1{\times}10^{-9}\)
& \(0.83 \pm 0.09\)
& \(1.1{\times}10^{-9} \pm 2.1{\times}10^{-9}\)
\\

&& \quad + ADSD
& \(0.18 \pm 0.10\)
& \(1.4{\times}10^{-11} \pm 2.4{\times}10^{-10}\)
& \(0.81 \pm 0.08\)
&  \(4.5{\times}10^{-10} \pm 1.5{\times}10^{-9}\)
\\

\bottomrule
\vspace{-2.5em}
\end{tabular}}
}
\end{table*}

The second element is the paper's normalized first-order error, Eqs.~(5)--(6) in \citet{wachter2006implementation}, specifically
\begin{equation}
\widetilde E=\max\Big\{
\frac{\|\tilde r_{\text{stat}}\|_\infty}{s_d},\;
\|\tilde c_e\|_\infty,\;
\|\tilde c_i+\tilde\sigma\|_\infty,\;
\frac{\tilde{\text{comp}}_{\max}}{s_c}\Big\},
\label{eq:etilde}
\end{equation}
with
\[s_d=\frac{\max\{s_{\max},(\|\tilde y\|_1+\|\tilde\pi\|_1+\|\tilde z_L\|_1+\|\tilde z_U\|_1)/(m+n)\}}{s_{\max}}\]
and $s_c=\max\{s_{\max},\|(\tilde\pi,\tilde z_L,\tilde z_U)\|_1/n\}/s_{\max}$,
$s_{\max}=100$. $\widetilde E$ replaces \texttt{meas} in best-state selection and stall detection only. Every acceptance gate and every reported residual continues to be evaluated in exact checker units through the map-back. Together, row scaling and a reweighted KKT progress measure help the solver continue through temporary increases in error while preserving the original convergence checks. The resulting solver significantly improves the pass rate, which demonstrates the combined method's effectiveness.

\section{Further Discussions}\label{app:exp}

We report the detailed numerical statistics for the generalization experiments in \S\ref{sec:exp-pf}. Tables~\ref{tab:topology-generalization} and \ref{tab:feasibility-generalization} provide the complete results, including running times and standard deviations that are omitted from Figure~\ref{f2}. For literature-space construction, we directly collect papers from Google Scholar and the Semantic Scholar API using domain-level keywords (e.g., ``Newton power flow convergence'' and ``stiff ODE integration''), resulting in the domain-specific literature database (e.g., 312 papers on PF equation solving and 376 papers on AC-OPF control).

Taking PF$\Delta$ as an example, weprovide some further discussions to help the general readers understand the impact of lower solver error from the perspective of professional power grid engineers.

\vspace{-0.5em}
\paragraph{On the practical essence of solver residue.} Specifically, it provides three practical impacts in realistic large-scale grids: (i) tighter residual control improves the numerical reliability of voltage and thermal-limit assessments following \texttt{N1} or \texttt{N2} contingencies. Near a regular operating point, the error in a monitored quantity $g$ satisfies the first-order relation $\delta g \approx \nabla g^\top J^{-1}r$, where $J$ is the power-flow Jacobian and $r$ is the residual; a smaller residual therefore improves the corresponding sensitivity-weighted error bound, enabling more reliable assessment of small security margins \citep{higham2002accuracy}. 

Second, lower residuals help distinguish small changes caused by redispatch, reactive-power support, or transformer adjustments from numerical solution errors. This is particularly valuable for finite-difference sensitivity calculations, in which errors from individual power-flow solutions are amplified by the inverse perturbation size \citep{more2012estimating}. 

Third, tighter power-balance satisfaction provides more consistent equilibrium initial conditions for dynamic simulations, reducing the potential for artificial transients caused by initialization mismatches \citep{molzahn2012initializing}. These benefits make residual control relevant beyond a binary convergence test, particularly under stressed or ill-conditioned operating conditions \citep{canizares1993point}; the required accuracy should be determined by downstream sensitivities and decision margins.

\vspace{-0.5em}
\paragraph{On the role of diagnosis.} Effective solver improvement requires connecting observed failures to the numerical methods capable of addressing them. Our case studies illustrate why diagnosis is therefore essential, and why simply exposing an agent directly to a broad literature space and asking it to improve the solver is often ineffective. In Appendix~\ref{app:qwen}, the baseline already implements a primal-dual interior-point method, but omits the specific scaling and progress-measure mechanisms needed to resolve its stalling behavior. Diagnosis converts this observed failure into a targeted search for row scaling and normalized KKT progress measures from \citet{wachter2006implementation}, well-established techniques underlying Ipopt and described in a classic work that should be already represented in the training data of frontier models. This suggests that access to numerical knowledge itself is not the primary barrier; rather, the central challenge is identifying which knowledge is relevant to the failure at hand and retrieving it accordingly.

In Appendix~\ref{app:gpt-pf}, diagnosis similarly identifies progress per Jacobian factorization as the relevant search objective, narrowing 312 papers to 82 candidates and ultimately six papers for full-text examination. Separately, the five-round GOC-500 experiment shows that repeated solver revisions alone do not close the gap to ADSD. Together, these observations support diagnosis as the mechanism that makes existing numerical knowledge actionable: it determines which methods to seek, explains why they are appropriate for the observed failure, and specifies the properties that their implementation must preserve.

\end{document}